\documentclass[runningheads]{llncs}

\usepackage[T1]{fontenc}
\usepackage{mathrsfs} 
\usepackage{epsfig}
\usepackage{epstopdf}
\usepackage{amsmath} 
\usepackage{multirow}
\usepackage{colortbl}
\usepackage{hyperref}
\usepackage[table]{xcolor}
\usepackage{booktabs} 
\usepackage{amssymb}
\usepackage{graphicx}

\title{CMEdataset: Advancing China Map Detection and Standardization with Digital Image Resources}
\author{
    Yan Xu\inst{1,2} \and
    Zhenqiang Zhang\inst{1,2} \and
    Zhiwei Zhou\inst{1,2} \and
    Liting Geng\inst{1,2} \and
    Yue Liu\inst{1,2} \and
    Jintao Li\inst{1,2,}\thanks{Corresponding author.}
}

\institute{
    Key Laboratory of Computing Power Network and Information Security, Ministry of Education, Shandong Computer Science Center (National Supercomputer Center in Jinan), Qilu University of Technology (Shandong Academy of Sciences), Jinan, China \\
    \and
    Shandong Provincial Key Laboratory of Computer Networks, Shandong Fundamental Research Center for Computer Science, Jinan, China
}

\date{March 2025}

\begin{document}

\maketitle 

\section{Introduction}
The digital images of China's maps play a crucial role in the field of map detection, especially concerning the global coverage of Chinese map datasets\cite{Li}. These digital images facilitate the accurate identification of China's national and provincial boundaries, contributing significantly to safeguarding national sovereignty, territorial integrity, and administrative management\cite{GaoJun2021}. Moreover, digital images are essential for map compliance detection, ensuring that maps adhere to regulations by correctly labeling place names and including key geographical elements, thereby maintaining the legality and standardization of maps. Additionally, digital images provide a rich data resource for the innovation and advancement of map detection technologies, driving the application of image processing, pattern recognition, and artificial intelligence\cite{Chen}. In certain cases, digital images also help prevent maps from containing information that may compromise national security or disclose classified information. In summary, digital images of China's maps play a vital role in ensuring map quality, compliance, security, and technological advancement.

Digital images play a significant role in map detection, particularly in accurately identifying China's national and provincial boundaries, ensuring the precision of geographical information. This is essential for maintaining national sovereignty, territorial integrity, and administrative governance\cite{mendel2016global}. Furthermore, digital images are critical in map compliance detection, as they ensure adherence to regulations by correctly labeling place names and preserving complete geographical elements, thereby safeguarding the legality and standardization of maps. More importantly, digital images provide a vast resource for the advancement of map detection technologies, fostering the application of image processing, pattern recognition, and artificial intelligence in this field\cite{Tang}. In certain contexts, digital images also help detect and prevent the inclusion of sensitive information in maps that could compromise national security or disclose classified details. In conclusion, China's map digital images play a crucial role in ensuring map quality, compliance, security, and the advancement of related technologies.

Currently, there is no publicly available dataset that specifically covers the five key aspects of ``problematic maps". Existing datasets mainly focus on general map data and lack detailed annotations and analyses of problematic maps, particularly in detecting national boundary misrepresentations, missing elements, blurred boundaries, incorrect labeling, and map compliance issues. Since these problematic maps often contain complex geographical errors and anomalies, existing public datasets are inadequate for effectively identifying and correcting these issues. Furthermore, the field of problematic map detection requires in-depth exploration of specific geographic areas, details, and error types. However, current public datasets suffer from limitations in sample diversity and coverage, making them insufficient for handling complex map anomalies and errors. The absence of a dedicated dataset that aligns with the five key areas restricts the application and development of map detection technologies in these specific domains.

Therefore, the creation of this dataset aims to provide diverse samples of problematic maps to support research and development in problematic map detection. By encompassing various types of map anomalies, such as misrepresented national boundaries, omitted islands, and blurred borders, the dataset ensures that detection algorithms can adapt to different map styles and effectively identify various issues. Additionally, the dataset supports high-precision map compliance detection, ensuring that maps adhere to relevant cartographic standards while enabling the automatic detection of errors to enhance map data quality and timeliness. This dataset not only provides extensive training data for problematic map detection technologies but also serves as a crucial resource for improving map compliance, national security monitoring, and map updates and maintenance. The application scenarios include map compliance detection and regulation, map updates and maintenance, national security and sensitive map monitoring, as well as fostering academic research and innovation in related technologies. Through this dataset, the development and application of problematic map detection technology can be effectively advanced, thereby enhancing the reliability and security of map data.

\section{Data Collection}
The digital images of China's maps in this dataset are sourced from the National Geographic Information Public Service Platform, as well as several publicly available online map platforms such as iStock, Geology, and others. These images primarily cover the topographic maps of China's entire geographical area. In addition, the dataset includes images based on the digitization of paper maps and high-definition digital maps generated through GIS technology. The time range of the images spans from 2015 to the present, covering various versions and update cycles of China's maps to ensure the timeliness, comprehensiveness, and completeness of the data.

The criteria for selecting the images are based on the following aspects:

\textbf{(1)} We have selected map types, including political maps, topographic maps, and others, to ensure the coverage of global maps within China's territory, with particular emphasis on choosing maps with rich colors. We also set a high resolution requirement for the collected map images, typically above 300dpi, to ensure the clarity of details and high-quality geographic information presentation. Images with lower resolution are excluded to ensure that the dataset meets the precise map detection needs.

\textbf{(2)} Currently, this dataset contains 1,455 digital images of China's maps, covering various types of maps, including but not limited to political and administrative maps, topographic maps, and hydrological maps, ensuring the diversity and representativeness of the data. Considering the research needs, the collected images clearly display important geographical features such as national borders and islands. This dataset provides rich geographical information and diverse map styles for the detection of problematic maps, ensuring its wide application potential.

\textbf{(3)} We have collected China's map images from multiple public platforms and data sources, making the data sources extensive and the image formats diverse. To ensure the consistency and standardization of the data, all collected map images have undergone unified format conversion and preprocessing, including standardizing image dimensions, resolution, and annotation formats, ensuring the reliability of the data for subsequent analysis and applications. However, due to the existence of different versions of map images and varied annotation standards, the accuracy of the annotations may have some discrepancies. To address this, the annotation work for the dataset has been undertaken by an experienced professional team and has undergone multiple rounds of verification and review to ensure the accuracy and consistency of the annotations. In addition, we have referred to several standardized map annotation systems and consulted with experts in the field to ensure the scientific and compliant nature of the annotations, further enhancing the quality of the dataset. At the same time, we have paid special attention to and actively collected common errors in map images, such as boundary misrepresentations, missing features, and other issues, and have meticulously labeled and categorized these errors during the annotation process. This ensures that the dataset can effectively support research in the detection and correction of problematic maps, further advancing the improvement of map data quality.

\section{Data Preprocessing}
In the process of constructing the dataset, image preprocessing, augmentation, and annotation are key steps to ensure data quality\cite{Huang}. First, in terms of image preprocessing, we performed denoising, cropping, and size standardization on all collected map images\cite{Yan}. The denoising operation effectively removed interference factors from the images, improving their clarity and quality. The cropping step ensured that key areas of the images were preserved while irrelevant parts were removed, thereby reducing redundancy in the computation. Size standardization unified the image size and resolution, providing consistent data input for subsequent model training\cite{2021,Zhao}.

In the image augmentation stage, we employed various data augmentation techniques, including rotation, translation, scaling, and cropping. These methods, by simulating different perspectives and spatial transformations, effectively increased the diversity of the training data, improving the model's generalization ability and allowing it to better adapt to map images in various real-world scenarios. By varying the angles and scales, the model's adaptability to image features was enhanced, thus improving its robustness and accuracy.

To ensure the quality of the annotations, we used professional annotation tools and established a strict annotation process and set of standards. During the annotation process, we annotated data in both YOLO and COCO formats according to different requirements. These two annotation formats satisfy the needs of different model training, increasing the dataset's versatility and flexibility. All annotation results underwent multiple rounds of validation and correction, employing a cross-validation mechanism where multiple annotators annotated the same image. The results were then compared to identify and correct inconsistencies, eliminating human errors. Additionally, we conducted regular annotation reviews, using random sampling and expert evaluations to ensure the accuracy and consistency of the annotations, promptly detecting and correcting deviations and errors in the annotation process.

Finally, after the dataset was constructed, we conducted a comprehensive evaluation of the quality of the preprocessed data. Through checks on the image quality, annotation accuracy, and the effects of data augmentation, we ensured the high quality and reliability of the dataset, providing a solid foundation for subsequent model training and research. This series of rigorous steps and methods not only enhanced the professionalism of the dataset but also provided reliable resources for related research and applications.

\section{Dataset Structure and Annotation Format}
During the construction of the dataset, we designed a clear and efficient data structure to facilitate data management and subsequent model use. Below are the detailed descriptions of the dataset structure and annotation format:
\begin{figure*}[h]
    \centering
    \includegraphics[width=12cm]{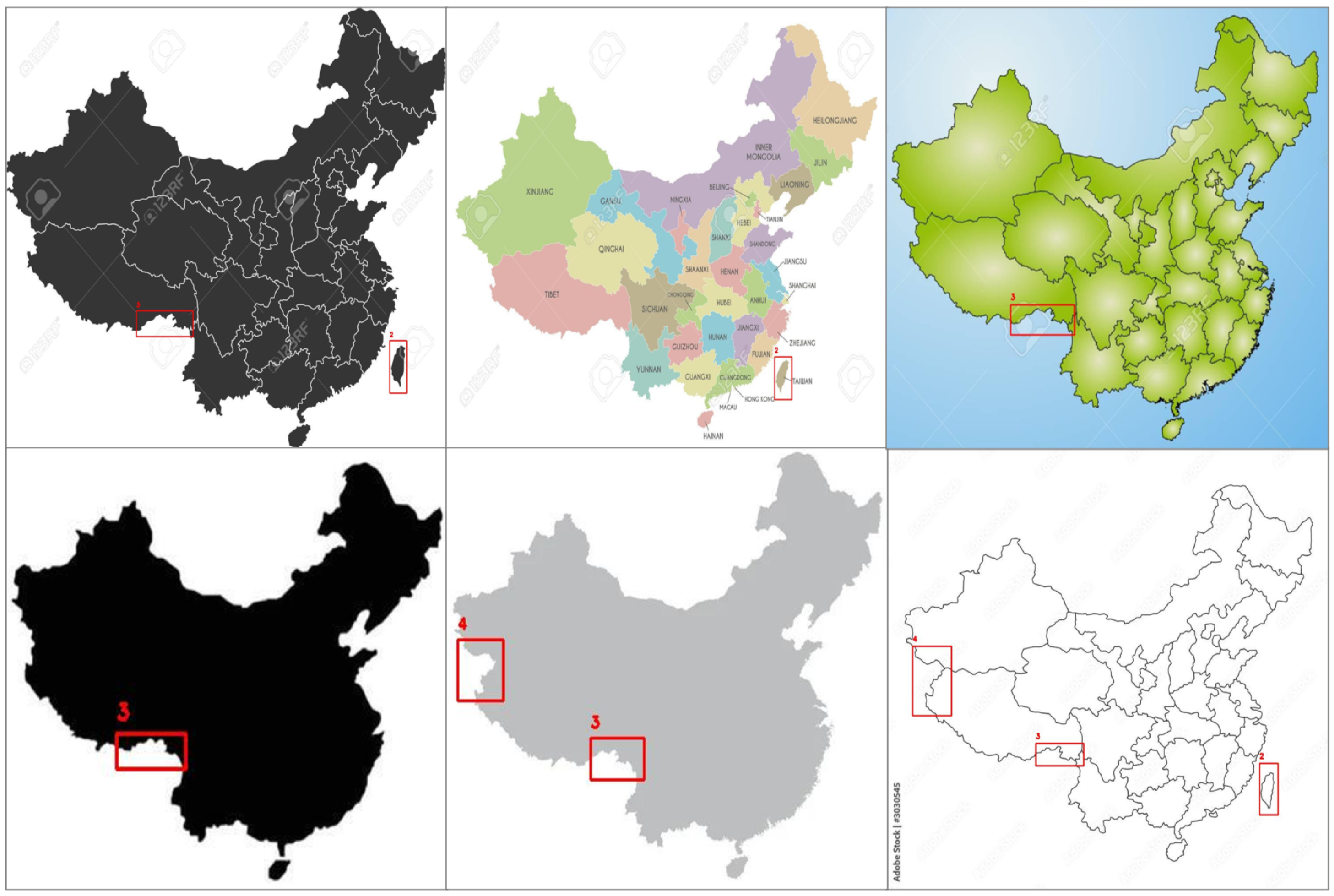}
    \caption{The figure displays six visualization results of the dataset annotations.}
    \label{fig:duibi}
\end{figure*}

\subsection{Dataset Structure}
The folder structure of the dataset adopts a hierarchical management system, mainly consisting of two parts: the image folder and the annotation folder.

\textbf{Image Folder:} This folder stores all the collected digital images of China’s maps. Each image file is stored according to a standard naming convention, ensuring that the images are ordered and easily retrievable.

\textbf{Annotation Folder:} This folder contains the annotation files corresponding to the image files. The annotation information is stored in both YOLO and COCO formats. Each image has a corresponding annotation file, and the filename matches the image file for easy pairing.

\subsection{Annotation Format}
During annotation, we used two common formats: YOLO format\cite{gao2024yolo} and COCO format\cite{Chen Li 2023mmship}, which meet different task requirements.

\begin{itemize}
    \item \textbf{YOLO Format:} The YOLO annotation file corresponding to each image is stored in .txt format. The contents of the file are recorded line by line for each target. Each line includes the class ID of the target, the coordinates of the center of the bounding box, and its width and height. All coordinates are normalized. The specific format is as follows:
    \[
    <class\_id> <x\_center> <y\_center> <width> <height>
    \]
    where <class\_id> is the class ID of the target (starting from 0), <x\_center> and <y\_center> are the coordinates of the center of the bounding box, and <width> and <height> are the width and height of the bounding box. These values are all normalized within the width and height of the image.

    \item \textbf{COCO Format:} The COCO annotation file corresponding to each image is stored in .json format, following the COCO standard annotation structure. The file includes several fields to describe the target information in the image. The main fields include:
    \begin{itemize}
        \item \texttt{images}: Contains basic information about the image, such as image ID, filename, etc.
        \item \texttt{annotations}: Includes detailed annotation information for all targets, with fields such as target ID, category ID, bounding box (bbox), segmentation information (segmentation), etc.
        \item \texttt{categories}: Defines the target categories and their corresponding category IDs.
    \end{itemize}
    Each target is annotated by recording its properties in a dictionary, as shown in the following example:
\end{itemize}
\begin{verbatim}
{
  "image_id": 1,
  "category_id": 2,
  "bbox": [x_min, y_min, width, height],
  "segmentation": [[...]],
  "area": area,
  "iscrowd": 0
}
\end{verbatim}
\subsection{Dataset Example}
\begin{table}[h]
    \caption{Overview of the target category and quantity distribution in the problematic map dataset}
    \label{tab:shujuji}
    \begin{tabular*}{\columnwidth}{@{\extracolsep\fill}ccccc}
        \hline\hline\noalign{\smallskip}
        ID&Category & Training & Test & Total  \\
        \midrule
        0 & South China Sea Islands &386& 86 & 472  \\
        1 & Diaoyu Island and Chiwei Islet &  331 &78 & 409  \\
        2 & Taiwan & 871 &211 & 1082  \\
        3 & Misaligned painting in southern Tibet & 681 &175 & 856 \\
        4 & Aksai Chin Incorrectly Depicted & 158 &43& 201  \\
        Total & — & \textbf{2427} &\textbf{593} & \textbf{3020} \\
        \hline
    \end{tabular*}
\end{table}
Table~\ref{tab:shujuji} provides a detailed distribution of the dataset, which is used for training and testing models to detect problematic regions in maps.

The detailed data distribution is shown below:
\begin{itemize}
    \item South China Sea Islands: The training set contains 386 samples, the test set contains 86 samples, totaling 472 samples.
    \item Diaoyu Islands and Chiwei Islet: The training set contains 331 samples, the test set contains 78 samples, totaling 409 samples.
    \item Taiwan: The training set contains 871 samples, the test set contains 211 samples, totaling 1882 samples.
    \item Southern Tibet Incorrectly Depicted: The training set contains 681 samples, the test set contains 175 samples, totaling 856 samples.
    \item Aksai Chin Incorrectly Depicted: The training set contains 158 samples, the test set contains 43 samples, totaling 201 samples.
\end{itemize}
In total, the training set contains \textbf{2427} samples, the test set contains \textbf{593} samples, and the entire dataset contains \textbf{3020} samples. 

This table provides detailed distribution information of the dataset for subsequent researchers, which helps to understand the data foundation during model training and evaluation.

In this chapter, we also present some examples from the dataset (as shown in Fig.~\ref{fig:duibi}) to demonstrate the results of our work.

\section{Dataset Evaluation and Experiments}
After completing the dataset construction, we conducted a comprehensive evaluation and experiments, focusing on the accuracy of annotations, model performance in different tasks, and a comparative analysis with existing datasets, in order to verify the quality and practicality of the dataset.
\begin{table}[h]
    \caption{The comparison table presents the performance evaluation metrics of our model against six other mainstream models on the CME dataset.}
    \label{tab:mo}
    \begin{tabular*}{\columnwidth}{@{\extracolsep\fill}ccccc}
          \hline\hline\noalign{\smallskip}
        Method & mAP@.5 & mAP@.5:.95 & Params & GFLOPs\\ 
       SMCA-DETR& 76.3\% & 34.4\%& 41M & 152\\
      Lite-DETR & 87.2\%& 46.9\%& 47M & 149 \\
        YOLOV9s   & 83.1\% & 48.2\% & 9.7M & 39.6\\
        YOLOV10s& 79.7\% & 46.7\% &8.0M & 24.8 \\
        YOLOV11s & 84.6\% & 48.5\%& 2.5M & 6.3 \\
        \hline
    \end{tabular*}
\end{table}
\subsection{Annotation Accuracy Evaluation}
To ensure the accuracy and consistency of the annotations, we employed multiple methods to evaluate and verify the annotation results. During the annotation process, we used cross-validation to check the annotation results. By assigning the annotation task to different annotators and comparing the annotation results of the same image, we ensured consistency and completeness of the annotations. Additionally, we wrote automated scripts to batch check parameters such as categories, bounding box sizes, and positions in the annotation files to ensure proper formatting and that the annotated data matches the image data. Through these multiple evaluation methods, we effectively reduced human errors and annotation bias during the annotation process, ensuring high-quality and consistent annotations.

\subsection{Target Detection Experiment}
In the target detection experiment, we used mainstream object detection algorithms (such as YOLO, DETR, etc.) to train and test on the dataset. The experimental results show that our constructed dataset excels in boundary detection, category recognition, and localization accuracy.

In the map error detection task, we used YOLOv9s~\cite{pan2024optimization}, YOLOv10s\cite{wang2024yolov10}, YOLOv11s\cite{khanam2024yolov11}, as well as Transformer-based Lite-DETR\cite{li2023lite} and SMCA-DETR\cite{gao2021fast} models to identify and correct issues such as boundary misrepresentation, target omissions, and boundary blurring. The experimental results demonstrated that the model achieved good performance in metrics like mAP@0.5 and mAP@0.5:0.95, especially excelling in the detection of small targets and blurred boundary regions.
\begin{figure*}[h]
    \centering
    \includegraphics[width=12cm]{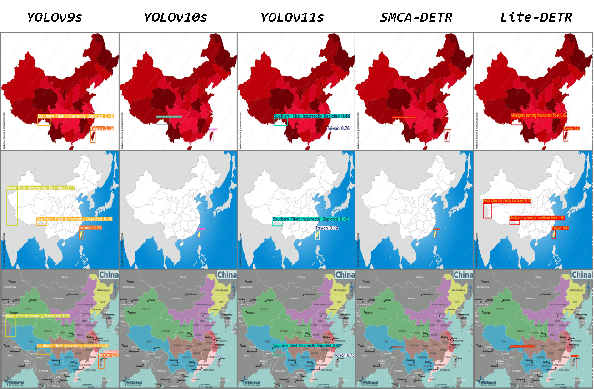}
    \caption{The figure illustrates the visualization results of our CME dataset across five of the most widely used object detection models. }
    \label{fig:mo}
\end{figure*}
The experimental results are shown in Table ~\ref{tab:mo} , demonstrating that this dataset effectively supports target detection and error localization tasks for complex map images, enhancing the model's adaptability and generalization ability for different types of map images.

The visualization results of the CME dataset in the model are shown in Fig.~\ref{fig:mo}. Although some models experienced missed detections, the overall performance of the dataset in real-world tasks is highly promising. These results demonstrate that the CME dataset is well-suited for complex map image detection and error localization tasks. The missed detections observed can be attributed to factors such as the inherent complexity of small-scale features, unclear boundaries, and overlapping or occluded objects, which are common challenges in map-related tasks. These aspects provide valuable insights for future work, where model improvements and the integration of advanced techniques, such as multi-scale feature fusion and attention mechanisms, can further enhance detection performance in more challenging areas. 

\section{Dataset Openness and Sharing}
This dataset has been publicly released on Google Drive, and researchers can directly access it through the  \href{https://drive.google.com/file/d/185iK7cG4J1Dws1wX1sUqVOt57CWwXnEG/view?usp=drive_link}{shared link} for convenient downloading and use. The attachment of the dataset includes a detailed description of the image sources and usage permissions. The dataset is available free of charge for scientific and academic research, and researchers are encouraged to cite and utilize it in their related studies to promote the development of the problematic map detection field. Future plans include further expanding the scale of the dataset by incorporating more types of maps and regions, particularly enhancing the complexity of boundaries and the diversity of targets, to more comprehensively cover various scenarios of problematic maps and further improve the model's generalization performance.

\section {Conclusion}
The CME dataset we have created fills a gap in map datasets, particularly by providing more representative data for areas with boundary errors, omissions, and ambiguities. By utilizing YOLO and COCO formats, the dataset has been made compatible with various detection models and frameworks, enhancing its versatility and broad applicability. Moreover, considering the unique characteristics of map data, our dataset includes annotations for small targets (such as islands) and complex boundary issues, which hold high detection value.

This dataset provides a foundational resource for the study of map problem detection and is expected to advance the development of automated map data detection technology, especially in precisely identifying boundary errors, omissions, and ambiguous areas on maps. In the future, the dataset could be applied to a wider range of scenarios, such as map repair, Geographic Information System (GIS) data processing, and automated map auditing, offering strong support for technological progress in related fields.

Looking ahead, we plan to further expand the dataset, particularly by adding more types of map data and problem areas to enhance its diversity and representativeness. At the same time, we will continue to improve the quality of data annotation and further strengthen the dataset's applicability in tasks involving complex backgrounds and small target detection. Additionally, we will explore more deep learning models based on this dataset to drive innovation in automated map detection technology and enhance the performance of algorithms in practical applications.

\bibliographystyle{splncs04}
\bibliography{3}

\end{document}